\documentclass[sn-mathphys-ay]{sn-jnl}
 
\usepackage{graphicx}%
\usepackage{multirow}%
\usepackage{amsmath,amssymb,amsfonts}%
\usepackage{amsthm}%
\usepackage{float}
\usepackage[title]{appendix}%
\usepackage{textcomp}%
\usepackage{manyfoot}%
\usepackage{booktabs}%
\usepackage{algorithm}%
\usepackage{algorithmicx}%
\usepackage{algpseudocode}%
\usepackage{listings}%
\usepackage{booktabs}

\begin{document}

\title[Article Title]{Efficient Training of Deep Networks using Guided Spectral Data Selection: A Step Toward Learning What You Need }

\author[1]{\fnm{Mohammadreza} \sur{Sharifi}}\email{sharifi.mohammadreza@mail.um.ac.ir}
\author*[1]{\fnm{Ahad} \sur{Harati}}\email{a.harati@um.ac.ir}



\affil*[1]{\orgdiv{Department of Computer Engineering}, \orgname{Ferdowsi University of Mashhad}, \orgaddress{\street{Azadi Square}, \city{Mashhad}, \postcode{9177948974}, \state{Khorasan-Razavi}, \country{Iran}}}




\abstract{
    
Effective data curation is essential for optimizing neural network training. In this paper, we present the Guided Spectrally Tuned Data Selection (GSTDS) algorithm, which dynamically adjusts the subset of data points used for training using an off-the-shelf pre-trained reference model. Based on a pre-scheduled filtering ratio, GSTDS effectively reduces the number of data points processed per batch. The proposed method ensures an efficient selection of the most informative data points for training while avoiding redundant or less beneficial computations. Preserving data points in each batch is performed based on spectral analysis. A Fiedler vector-based scoring mechanism removes the filtered portion of the batch, lightening the resource requirements of the learning. The proposed data selection approach not only streamlines the training process but also promotes improved generalization and accuracy. Extensive experiments on standard image classification benchmarks, including CIFAR-10, Oxford-IIIT Pet, and Oxford-Flowers, demonstrate that GSTDS outperforms standard training scenarios and JEST, a recent state-of-the-art data curation method, on several key factors. It is shown that GSTDS achieves notable reductions in computational requirements, up to four times; without compromising performance. GSTDS exhibits a considerable growth in terms of accuracy under the limited computational resource usage in contrast to other methodologies.
These promising results underscore the potential of spectral-based data selection as a scalable solution for resource-efficient deep learning and motivate further exploration into adaptive data curation strategies. You can find the code at \burl{https://github.com/rezasharifi82/GSTDS}.

}
\maketitle



\section{Introduction}
The quality of training data is a cornerstone of deep learning model performance. While large-scale datasets have propelled advancements in natural language processing ~(\cite{brown2020language}), computer vision~(\cite{sun2017revisiting}), and multimodal learning~(\cite{radford2021learning,jia2021scaling}), not all data points contribute equally to model training. Datasets often contain redundant, irrelevant, or non-informative samples, which increase computational overhead without significantly enhancing learning outcomes. Identifying and selecting the most informative data points is therefore essential to improve the efficiency of the model, reduce resource consumption, and improve generalization capabilities~(\cite{kaplan2020scaling,sorscher2022beyond}).

Recent advancements in automated data curation, such as active learning, have demonstrated the potential of dynamically selecting valuable data points based on heuristic approaches, which can result in noticeable training efficiency improvement~(\cite{settles2009active,cohn1996active}). Spectral analysis, particularly through the use of the Fiedler vector, has emerged as a robust mechanism for scoring data points within batches. This approach captures the geometric relationships between samples, enabling a principled evaluation of their informativeness~(\cite{von2007tutorial,liu2018spectral}). Furthermore, recent studies have shown that processing data collectively at the batch level, rather than individually, significantly improves the assessment of relative informativeness, outperforming traditional methods~(\cite{evans2024data}).

A key challenge in data curation lies in balancing the rigor of sample selection with the model's accuracy. By integrating spectral scoring with a pre-scheduled filtering ratio, we can achieve precise control over computational costs while maintaining model performance. To ensure resource allocation matches the training process dynamics, we smoothly increase filtering ratio to effectively work with larger batch sizes and enhance final stages of learning ~(\cite{smith2017don}).

In this study, we introduce Guided Spectrally Tuned Data Selection (GSTDS), a novel approach that combines spectral analysis via a scoring mechanism based on the Fiedler vector, and a dynamically adjusted filtering ratio to optimize batch-level data selection. 
Our method leverages a reference model to extract features, serving as a blueprint for the training process and enhancing the overall acquisition of information. This methodology is further combined with a filtering ratio schedule, which can be employed as a predefined plan to elevate the effectiveness of the training dynamics.
The filtering ratio schedule is adapted to select the desired amount of input data, which, as a hyperparameter, specifies the available computational resources. A considerable increase in accuracy is observed, especially when the task is more complex (i.e., when facing a greater number of classes and fewer samples per class). In such settings, GSTDS provides an appropriate curriculum for learning, helping to adjust the difficulty of the training process in a suitable manner.

Extensive experiments across multiple datasets demonstrate that our approach achieves significant performance under the same resource consumption. For instance, on the Oxford-Flowers, our method achieves 35.05\% improvement in accuracy compared to standard training methodology. 


\section{Related Works}\label{Related}
\textbf{Batch level selection}.
Previous studies have shown that selecting data points at the batch level can significantly improve model performance and training efficiency ~(\cite{coleman2019selection, sachdeva2024train, evans2024data}).
 Using the overall structure of a batch in comparison to individual data points, the model can learn those enhancing data points rather than all of them. In other words, data points have different expressions when they are to be judged at batch level ~(\cite{evans2024data}).
Besides all relevant studies on active learning ~(\cite{sener2017active, csiba2018importance, settles2009active}), our study simply filters out unnecessary data points in a batch manner to save resources.

 \textbf{Data curation using spectral methods}.
According to previous studies ~(\cite{von2007tutorial, belkin2003laplacian}), we are able to use some spectral approaches to extract the most connected or the most informative elements within an arbitrary set.
Since the Fiedler vector had shown a promising performance on extracting
the appropriate elements ~(\cite{manguoglu2010highly, shaham2018spectralnet, zhao2007spectral}),
we used that in a meaningful manner to extract the most-informative data points in each batch.

 \textbf{Scheduled Filter Ratio}.  
Deterministic data retention schedules balance efficiency and performance in deep learning. Previous studies used fixed batch scaling ~(\cite{smith2017don}) or increasing subset sizes, treating the filter ratio as a static hyperparameter. Adaptive strategies like dynamic batch size adjustments ~(\cite{gao2020balancing}) and AdAdaGrad ~(\cite{lau2024adadagrad}) enhance training efficiency and generalization.
In contrast, our approach utilizes a pre-computed, time-variant schedule to adjust the filter ratio throughout training .This dynamic schedule, informed by spectral characteristics like Fiedler vector scores, progressively adjusts the filter ratio to mimic a curriculum learning ~(\cite{bengio2009curriculum}) strategy, unlike prior static ratio methods.

 \textbf{Ratio-Based Curriculum Learning}.
Curriculum learning is a deep learning strategy that enhances training by presenting data in a meaningful order, often progressing from simpler to more complex examples ~(\cite{bengio2009curriculum,wang2021survey,soviany2022curriculum}). We employ a scheduled filter ratio, guided by spectral analysis and specifically Fiedler vector scores, to implicitly create a curriculum effect. This time-variant filter dynamically adjusts the proportion of training data used per epoch, prioritizing data subsets identified by Fiedler scores as spectrally informative. By selectively focusing on these data through a deterministic schedule, our ratio-based method aims to facilitate a more effective and focused learning process, without directly altering the inherent difficulty of individual data points.

 \textbf{Reference Model-Guided Active Learning}.  
Numerous strategies have been proposed for active learning, spanning both online and offline paradigms. Recent studies have shown that using a reference model to provide a comprehensive dataset overview can serve as a blueprint for selecting the most informative samples ~(\cite{settles2009active, gal2017deep, evans2024data}). This approach not only enhances model performance but also streamlines the decision-making process for prioritizing data points for annotation. In our study, we adopted a similar strategy by leveraging feature vectors extracted from each data point.



\section{Methods}\label{Methods}
In this study, we present a structured procedure designed to effectively utilize informative data, optimizing the training process and improving overall model efficiency. Initially, features are extracted from the entire, indexed dataset using a pre-trained model.  Subsequently, a time-variant filter ratio schedule is computed that constrains data point selection from each batch.  Further, using feature guidance from the pre-trained model, a Fiedler vector-based spectral score is employed to identify and select maximally informative data points, while adhering to the filter ratio's data budget. The methodological components of this procedure are detailed in the subsequent subsections.

\subsection{Feature Extraction and Model Architecture}\label{Feature}
Feature vectors for input images were extracted using a pre-trained ResNet-50 model ~(\cite{resnet}), a deep convolutional neural network initially trained on the ImageNet dataset ~(\cite{deng2009imagenet}). Specifically, the activations from the penultimate layer of ResNet-50 were employed as the feature representation for each data point. We call this the \emph{Reference Model} in this study, as it provides a comprehensive overview of the data points and serves as a blueprint for selecting the most informative samples. The Reference Model layers are frozen and the corresponding weights did not update during the feature extraction phase. meanwhile, we used ResNet-18 (pre-trained) ~(\cite{resnet}) as the main learner model for all experiments.

\subsection{ Scheduled Filter-Ratio sequence} \label{curriculum_filter}
In this study, the filter ratio $\mathcal{F}_i$ denotes the proportion of data points from batch $i$ that are input to the model during the current epoch. A higher ratio indicates that a greater number of data points from this batch are going to be processed by the model. The filter ratio for each batch is associated with a global sequence $\mathcal{G}$, defined as the collection of all filter ratios constructed based on the total number of batches processed during the training process. Specifically, if there are $n$ epochs and $m$ batches per epoch, then $\mathcal{G}$ contains $n \times m$ elements. The policy governing this sequence is highly variable and depends on the specific application being considered.

We used optimized sigmoid function $\mathcal{S}$ which defined as below:

 \[
\forall x \in \mathcal{G}, \quad f(x) = a + \frac{b-a}{1 + e^{-k(x - x_0)}}
\]

Wheree:
\begin{itemize}
    \item \( a \) is the lower bound of the sigmoid,
    \item \( b \) is the upper bound of the sigmoid,
  \item \( k \) is the steepness of the function,
  \item \( x_0 \) is the midpoint of the sigmoid.
\end{itemize}

Consequently, the filter ratio for each batch is defined as:
\[
\mathcal{F}_i = f(\mathcal{G}_i)
\]
where $\mathcal{G}_i$ denotes the $i$-th element of the global sequence $\mathcal{G}$.

 The appropriate value corresponding to our customized function is simply optimized in order
mostly satisfy some constraints such as:
\[
\text{Let } f(x) \text{ satisfy }
\left\{
\begin{array}{l}
\displaystyle \lim_{T\to\infty}\frac{1}{T}\int_{0}^{T} f(x)\,dx = 0.3, \\[1em]
\displaystyle \max_{x}\,f(x) = 0.88, \\[1em]
\displaystyle \min_{x}\,f(x) = 0.18.
\end{array}
\right.
\]
This is what we can denote as a scheduled filter-ratio sequence, 
which is used to start with a lower filter ratio and gradually increase it to a higher filter ratio during the training process.

 \subsection{Data selection} \label{data_selection}
The data selection process is a pivotal aspect of our algorithm. To isolate the most informative data points within each batch, we employ a spectral analysis method that capitalizes on the interrelationships among data points, thereby prioritizing those that most effectively enhance the learning process. For a comprehensive understanding, we delineate several key components in the subsequent sections.

Let \(\mathcal{Q} = \{q_1, q_2, \dots, q_N\}\) denote the entire dataset, which is partitioned into \(M\) disjoint batches:
\[
\mathcal{Q} = \bigcup_{k=1}^{M} \mathcal{B}_k, \quad \text{with } \mathcal{B}_i \cap \mathcal{B}_j = \varnothing \text{ for } i \neq j.
\]

For each data point \(q_i \in \mathcal{Q}\), let \(\mathbf{v}_i \in \mathbb{R}^d\) be its extracted feature vector. Define the complete set of feature vectors as:
\[
\mathcal{V} = \{\mathbf{v}_1, \mathbf{v}_2, \dots, \mathbf{v}_N\}.
\]

For each batch \(\mathcal{B}_k\), let the corresponding set of feature vectors be given by:
\[
\mathcal{V}_k = \{\mathbf{v}_i \in \mathcal{V} \mid q_i \in \mathcal{B}_k\}.
\]

Using the feature vectors in \(\mathcal{V}_k\), the cosine similarity matrix \(S^{(k)} \in \mathbb{R}^{|\mathcal{V}_k| \times |\mathcal{V}_k|}\) is defined as:
\[
S^{(k)}_{ij} = \frac{\mathbf{v}_i \cdot \mathbf{v}_j}{\|\mathbf{v}_i\|\,\|\mathbf{v}_j\|}, \quad \text{for } \mathbf{v}_i, \mathbf{v}_j \in \mathcal{V}_k.
\]

The corresponding degree matrix \(D^{(k)} \in \mathbb{R}^{|\mathcal{V}_k| \times |\mathcal{V}_k|}\) is a diagonal matrix with entries:
\[
D^{(k)}_{ii} = \sum_{j} S^{(k)}_{ij}.
\]

Finally, the Laplacian matrix \(L^{(k)}\) for batch \(\mathcal{B}_k\) is given by:
\[
L^{(k)} = D^{(k)} - S^{(k)}.
\]

By extracting the \textbf{Fiedler vector}, which is the eigenvector corresponding to the second smallest eigenvalue of the Laplacian matrix \( L^{(k)} \), we can rank the data points based on their significance in defining the batch structure. This allows us to select the most informative samples from that batch for training.

Let \( \phi \in \mathbb{R}^N \) denote the Fiedler vector extracted from the Laplacian matrix \( L \). We define the ranked list of $\phi$ components as:

\[
\mathcal{R} = \{ \phi_{(1)}, \phi_{(2)}, \dotsc, \phi_{(N)} \},
\]

ordered such that:

\[
\phi_{(1)} \geq \phi_{(2)} \geq \dotsb \geq \phi_{(N)}.
\]

Here, \( \phi_{(1)} \) represents the most informative element, with the informativeness decreasing accordingly for subsequent elements.

As we defined before (Section~\ref{curriculum_filter}), the overall filter ratio policy is determined by the curriculum filter-ratio sequence for the whole training process at the beginning. Since we have \(\mathcal{F}_k\) for the current batch of the current epoch, we can easily calculate the number of data points that should be selected from that batch at this time.

Given a batch \( \mathcal{B}_i \), let \( |\mathcal{B}_i| \) denote the total number of data points in the batch. The filter ratio \( \mathcal{F}_k \) determines the proportion of data points to be selected. The total number of selected data points \( n_i \) is computed as:

\[
n_i = \lfloor \mathcal{F}_i \cdot |\mathcal{B}_i| \rfloor.
\]

The selection process is divided into three steps:

\begin{itemize}
    \item \textbf{Deterministic Top 50\% Selection}
The top \( \lfloor n_i / 2 \rfloor \) data points are selected based on their ranking in the Fiedler vector \( \phi_i \) (derived from the Laplacian matrix of the batch):

\[
\mathcal{U}_{i}^{(1)} = \{ q_j \mid q_j \in \mathcal{B}_i, \ j \in \text{Top}_{\lfloor n_i / 2 \rfloor} (\phi_i) \},
\]
where \( \text{Top}_{k} (\phi_i) \) represents the set of indices corresponding to the largest \( k \) values of \( \phi_i \).

\item \textbf{Weighted Random Sampling for Remaining 50\%}
To have better selection and avoid data leakage,
we assigned the refined form of the point-wise reference model loss to each data point correspondingly.

Let \( \ell(q_j) \) denote the point-wise loss of a data point \( q_j \) computed using the reference model. We define the weight of each data point as:

\[
w_j = \frac{1}{\ell(q_j) + \epsilon},
\]

where \( \epsilon \) is a small constant added for numerical stability. These weights are computed for all data points in the batch \( \mathcal{B}_i \) without normalization.

Let

\[
\mathcal{R}_i = \mathcal{B}_i \setminus \mathcal{U}_{i}^{(1)}
\]
be the set of data points not selected in the deterministic phase. For each \( q_j \in \mathcal{R}_i \), we normalize the weights to form a probability distribution:

\[
P(q_j) = \frac{w_j}{\sum_{q_k \in \mathcal{R}_i} w_k}.
\]

Then, we select \( \lfloor n_i / 2 \rfloor \) data points from \( \mathcal{R}_i \) using weighted random sampling:

\[
\mathcal{U}_{i}^{(2)} = \text{RandomSample}\left( \mathcal{R}_i, \lfloor n_i / 2 \rfloor, P \right).
\]

\item \textbf{Final Selection Set}

The final selection set for batch \( i \) is given by:

\[
\mathcal{U}_i = \mathcal{U}_{i}^{(1)} \cup \mathcal{U}_{i}^{(2)}.
\]

Since \( \mathcal{U}_{i}^{(1)} \) and \( \mathcal{U}_{i}^{(2)} \) are disjoint subsets, their total size is:

\[
|\mathcal{U}_i| = |\mathcal{U}_{i}^{(1)}| + |\mathcal{U}_{i}^{(2)}| = \lfloor n_i / 2 \rfloor + \lfloor n_i / 2 \rfloor = n_i.
\]

Thus, the total number of selected data points is precisely \( n_i \), as initially determined.

\end{itemize}

\begin{algorithm}[htbp] 
    \caption{Spectral Batch Selection with Curriculum Filtering}
    \begin{algorithmic}[1]
    \Require Dataset $\mathcal{Q}$, batch index $i$, total batches $M$, filter ratio $\mathcal{F}_i$
    \Ensure Selected subset $\mathcal{S}_i$
    \State Partition dataset: $\mathcal{B} = \{\mathcal{B}_1, \mathcal{B}_2, \dots, \mathcal{B}_M\}$
    \State Extract feature vectors: $\mathcal{V} = \{ v_1, v_2, \dots, v_N \}$
    \State Select batch features: $\mathcal{V}_i = \{ v_j \mid q_j \in \mathcal{B}_i \}$
    \State Compute similarity matrix: $S_i \gets \cos(\mathcal{V}_i, \mathcal{V}_i^\top)$
    \State Compute degree matrix: $D_i \gets \text{diag}(\sum S_i, \text{axis}=1)$
    \State Compute Laplacian matrix: $L_i \gets D_i - S_i$
    \State Extract Fiedler vector: $\phi_i \gets \text{eigenvector}_2(L_i)$
    \State Rank data points: $\mathcal{R}_i \gets \text{sort}(\phi_i, \text{descending})$
    \State Compute selection count: $n_i \gets \lfloor \mathcal{F}_i \cdot |\mathcal{B}_i| \rfloor$
    \State Select top 50\%: $\mathcal{S}_1 \gets \mathcal{R}_i[1 : \lfloor n_i / 2 \rfloor]$
    \State Select remaining 50\% via weighted random sampling:
    \State \hspace{1em} Compute weights: $w_j \gets |\phi_{i,j}| / \sum |\phi_{i,j}|$
    \State \hspace{1em} Sample indices: $\mathcal{S}_2 \gets \text{WeightedSample}(\mathcal{R}_i[\lfloor n_i / 2 \rfloor :], \lfloor n_i / 2 \rfloor, w)$
    \State Combine selected data points: $\mathcal{S}_i \gets \mathcal{S}_1 \cup \mathcal{S}_2$
    \State \Return Selected subset $\mathcal{S}_i$
    \end{algorithmic}
    \label{alg:spectral_batch}
\end{algorithm}

Algorithm~\ref{alg:spectral_batch} employs an exploration-exploitation strategy for data point selection, reminiscent of reinforcement learning approaches (\cite{sutton2018reinforcement}). It selects the top half of data points based on Fiedler scores (exploitation), and the remaining half via weighted random sampling (exploration), with weights inversely proportional to reference model point-wise losses. This balances the selection of high-confidence points with the discovery of potentially informative, high-loss regions.

By utilizing this approach in conjunction with a scheduled filter ratio, a curriculum learning procedure is implicitly leveraged.  Initially, in each batch, data points are strategically selected based on a dual criterion. This selective strategy focuses initial training on a data subset designed to facilitate effective early learning by presenting a less complex yet informative training scenario.  Subsequently, as the filter ratio schedule increases, the quantity and diversity of the training data are gradually expanded. This progressive data infusion systematically elevates the challenge, thereby fostering a stepwise and robust enhancement of the model's generalization capabilities, characteristic of a curriculum learning paradigm.


\section{Results}\label{results}
To evaluate the efficacy of our proposed approach, experiments were conducted on three benchmark image classification datasets: Oxford Flowers 102 ~(\cite{Nilsback08}), Oxford-IIIT Pet ~(\cite{parkhi12a}), and CIFAR-10 ~(\cite{krizhevsky2009learning}). These datasets, comprising diverse image characteristics and classification challenges, were selected to rigorously assess model performance.  Specifically, Oxford Flowers 102 (8,189 images, 102 categories) presents fine-grained classification challenges, Oxford-IIIT Pet (7,349 images, 37 pet breeds) evaluates object recognition, and CIFAR-10 (60,000 32x32 color images, 10 classes) benchmarks performance on low-resolution, high-variability images.  Performance comparisons were made against a standard training procedure using the full dataset ~(\cite{Goodfellow-et-al-2016,lecun2015deep}) and a uni-modal adaptation of Google DeepMind's JEST algorithm ~(\cite{evans2024data}), which is guided by a reference image classifier instead of the text data in the original work, for fair comparison. 

All datasets underwent identical preprocessing and augmentation pipelines including random resized cropping, horizontal flipping, color jittering, random rotation, and normalization; test sets remained unaugmented. Datasets were indexed and partitioned into training, validation, and test splits following default PyTorch/torchvision configurations ~(\cite{paszke2019pytorch, marcel2010torchvision}). Training was performed using SGD optimizer with a learning rate of $0.001$, and batch sizes of 64 for Oxford-Flowers and 128 for both Oxford-IIIT Pet and CIFAR-10.  The experiments were carried out on a system equipped with an NVIDIA GeForce RTX 3050 GPU, an Intel Core i5-14500 CPU, and 16GB of RAM, using the PyTorch deep learning framework ~(\cite{paszke2019pytorch}).

\subsection{Different schedules for adapting Filter-Ratio}

\begin{figure}[htbp]
        \centering
        \includegraphics[width=\textwidth]{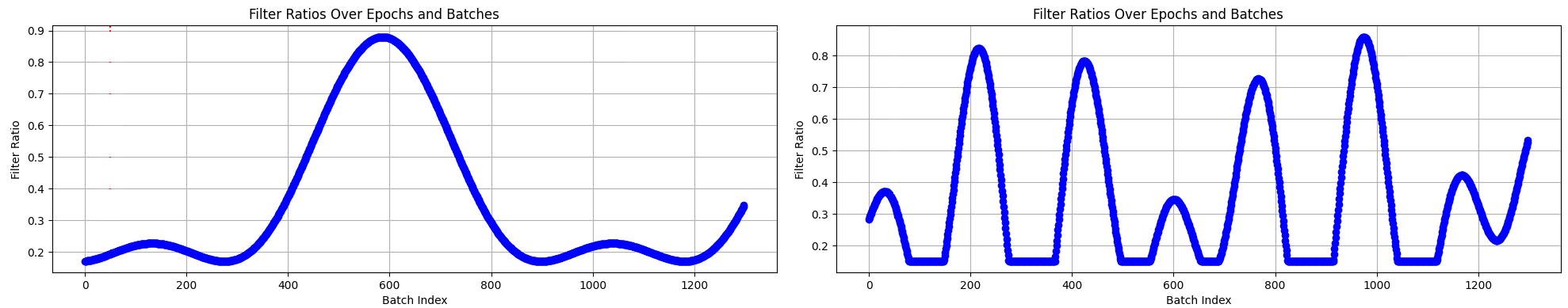}
        \caption{Oscillatory Filter-ratio policies. \textbf{Left:} Sinc function. \textbf{Right:} Lower-Bounded Sinusoid.}
        \label{fig:filter_ratio_policies1}
    \end{figure}

\begin{figure}[htbp]

        \centering
        \includegraphics[width=\textwidth]{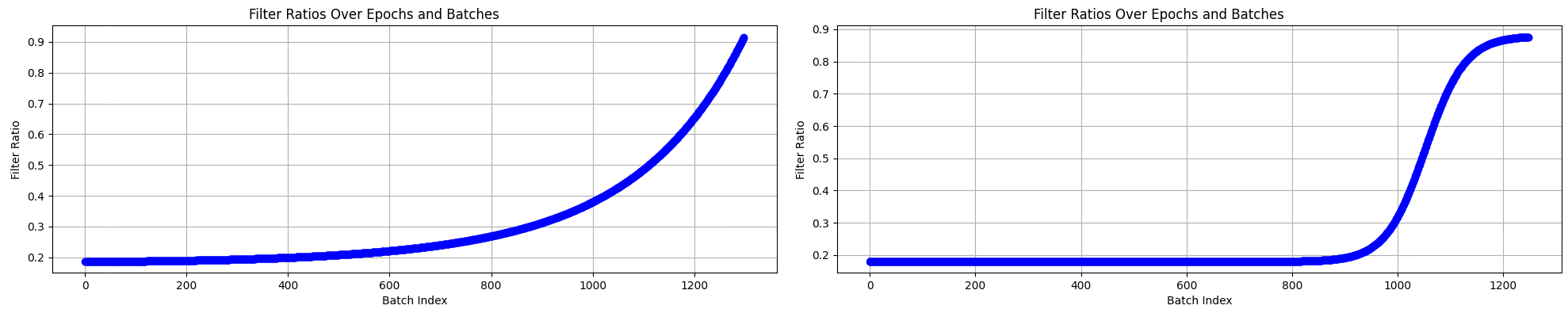}
        \caption{Non-linear Filter-ratio policies. \textbf{Left:} Gamma function. \textbf{Right:} Sigmoid.}
         \label{fig:filter_ratio_policies2}
    \end{figure}

\begin{figure}[htbp]
        \centering
        \includegraphics[width=\textwidth]{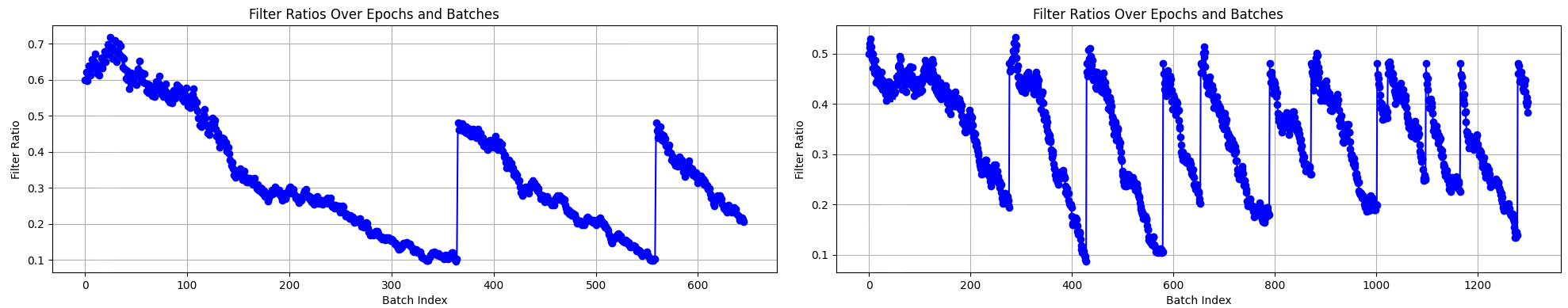}
        \caption{Filter-ratio sequences using AIMD-inspired policies. \textbf{Left:} Mellow AIMD policy. \textbf{Right:} Aggressive AIMD policy.}
        \label{fig:filter_ratio_policies3}
    \end{figure}

As established in Section~\ref{curriculum_filter}, the scheduled filter ratio constitutes a crucial component of the GSTDS algorithm. In this section, we present the experimental findings from our evaluation of diverse filter ratio scheduling policies, with a particular emphasis on results obtained using the Oxford-IIIT Pet dataset.
The filter ratio schedules implemented by different policies are visually represented in Figures~\ref{fig:filter_ratio_policies1}, ~\ref{fig:filter_ratio_policies2}, and ~\ref{fig:filter_ratio_policies3}.

    Using oscillatory policies such as bounded sinusoidal or sinc functions (shown in Figure\ref{fig:filter_ratio_policies1}) was thought to improve performance by reaching higher filter ratios during the middle of training, rather than at the end. However, the results show that this approach does not lead to a better result (shown in Table\ref{tab:performance_comparison_policies_hi}).
    
        As shown in Figure ~\ref{fig:filter_ratio_policies2}, adjusted-gamma and adjusted-sigmoid functions can control data usage over time. By using non-linear policies, higher filter ratios are shifted to later epochs, which, as shown in Table ~\ref{tab:performance_comparison_policies_hi}, leads to improved performance.

The filter-ratio policies depicted in Figure~\ref{fig:filter_ratio_policies3} are inspired by the Additive Increase Multiplicative Decrease (AIMD) mechanism (~\cite{Chiu1989AIMD}), commonly used for congestion control in computer networks. The \textit{Mellow AIMD} policy gradually reduces the filter ratio additively, but when the model's performance degrades, it first attempts to improve performance by increasing the ratio multiplicatively. If poor performance persists (due to some consecutive loss), the ratio jumps to an average value. Similarly, the \textit{Aggressive AIMD} policy adjusts the ratio more sharply, but it also suffers from high oscillations due to abrupt changes. These oscillations lead to instability during training. In contrast, the sigmoid policy demonstrates smoother transitions and better stability, effectively mitigating these issues. As shown in Table~\ref{tab:performance_comparison_policies_hi}, the sigmoid policy generally outperforms other variants. Due to its superior performance, we adopted the sigmoid policy as the pre-scheduled mechanism for filter ratio adjustment in subsequent experiments.

\begin{table}[htbp]
    \centering
    \small
    \caption{Performance Comparison on Oxford-IIIT Pet dataset}\label{tab:performance_comparison_policies_hi}
    \begin{tabular}{lcccccc}
    \toprule
    Metric                   & Sinc  & Sinu. & Gam.  & Mel.AIMD & Agg.AIMD & \textbf{Sig} \\
    \midrule
    Train Acc.               & 79.98 & 81.10    & \underline{87.78}  & 83.82    & 81.24    & \textbf{86.30} \\
    Val. Acc.                & 80.16 & 83.15    & 81.79  & \underline{82.88}    & 80.71    & \textbf{85.87} \\
    Test Acc.                & 87.98  & 87.62   & \underline{89.04}  & 87.27    & 88.11    & \textbf{89.69} \\
    Train Loss               & 0.6978 & 0.6277  & \underline{0.4920} & 0.6577   & 0.6600   & \textbf{0.4873} \\
    Val. Loss                & 0.6329 & \underline{0.5819}  & 0.6142 & 0.6215   & 0.6393   & \textbf{0.5790} \\
    Test Loss                & 0.3889 & 0.3934  & \underline{0.3698} & 0.4408   & 0.3816   & \textbf{0.3429} \\
    FLOPs$^{*}$              & 9.74   & 9.80    & \underline{8.69}   & 9.47     & 9.41    & \textbf{8.38} \\
    Data Usage(\%)$^{**}$    & 0.3473 & 0.3500  & \underline{0.3125} & 0.3263   & 0.3363   & \textbf{0.3012} \\
    \bottomrule
    \end{tabular}
    \begin{flushleft}
    \footnotesize
    $^{*}$ FLOPs (Floating Point Operations) are reported in $10^{13}$. Lower values are better, so the lowest is bolded, and the second-lowest is underlined. \\
    $^{**}$ The Avg.Dat. field represents the average amount of data fed to the model during training, measured by the average filter ratios over the entire process. \\
    \textit{Note: The best values in each row are bolded, while the second best values are underlined for easier comparison.}
    \end{flushleft}
\end{table}


\subsection{ Comparison of Accuracy and Efficiency for Different Filtering Methods}\label{performance}

\begin{figure}[htbp]
    \centering
    \includegraphics[width=\textwidth]{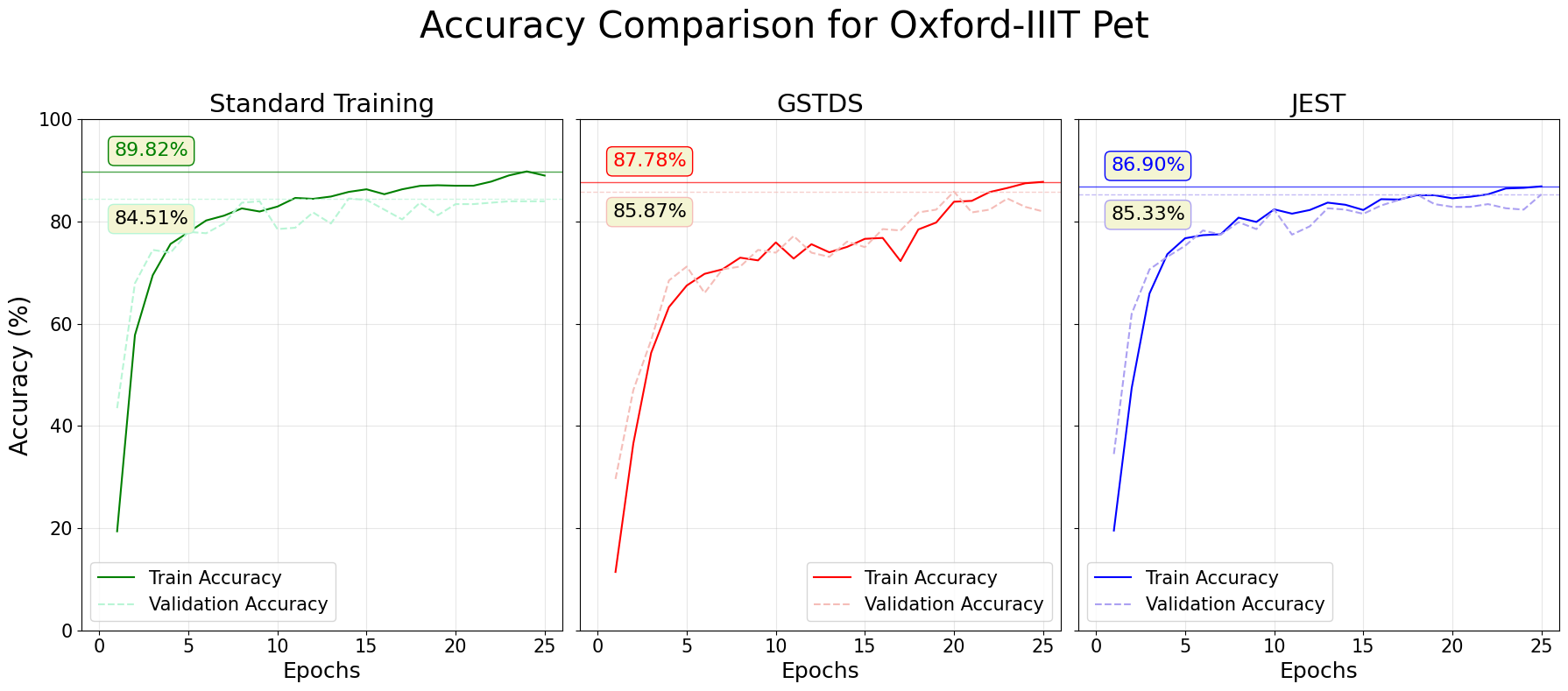}
    \caption{Train and Validation Curves of Oxford-IIIT Pet. From left to right: Standard Training, GSTDS, JEST.}\label{fig:Oxford_curves}
\end{figure}

\begin{figure}[htbp] 
    \centering
    \includegraphics[width=\textwidth]{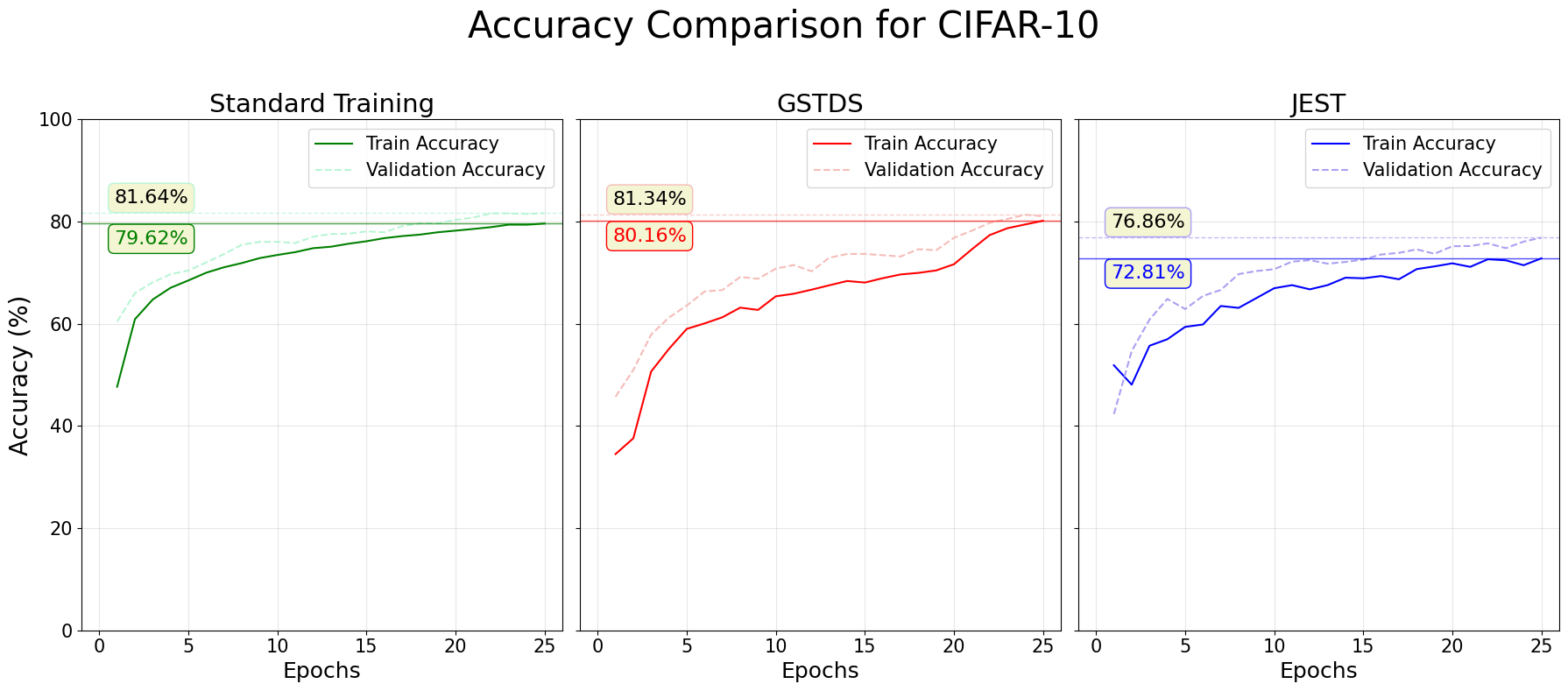}
    \caption{Train and Validation Curves of CIFAR-10. From left to right: Standard Training, GSTDS, JEST.}\label{fig:CIFAR_curves}
\end{figure}

\begin{figure}[htbp] 
    \centering
    \includegraphics[width=\textwidth]{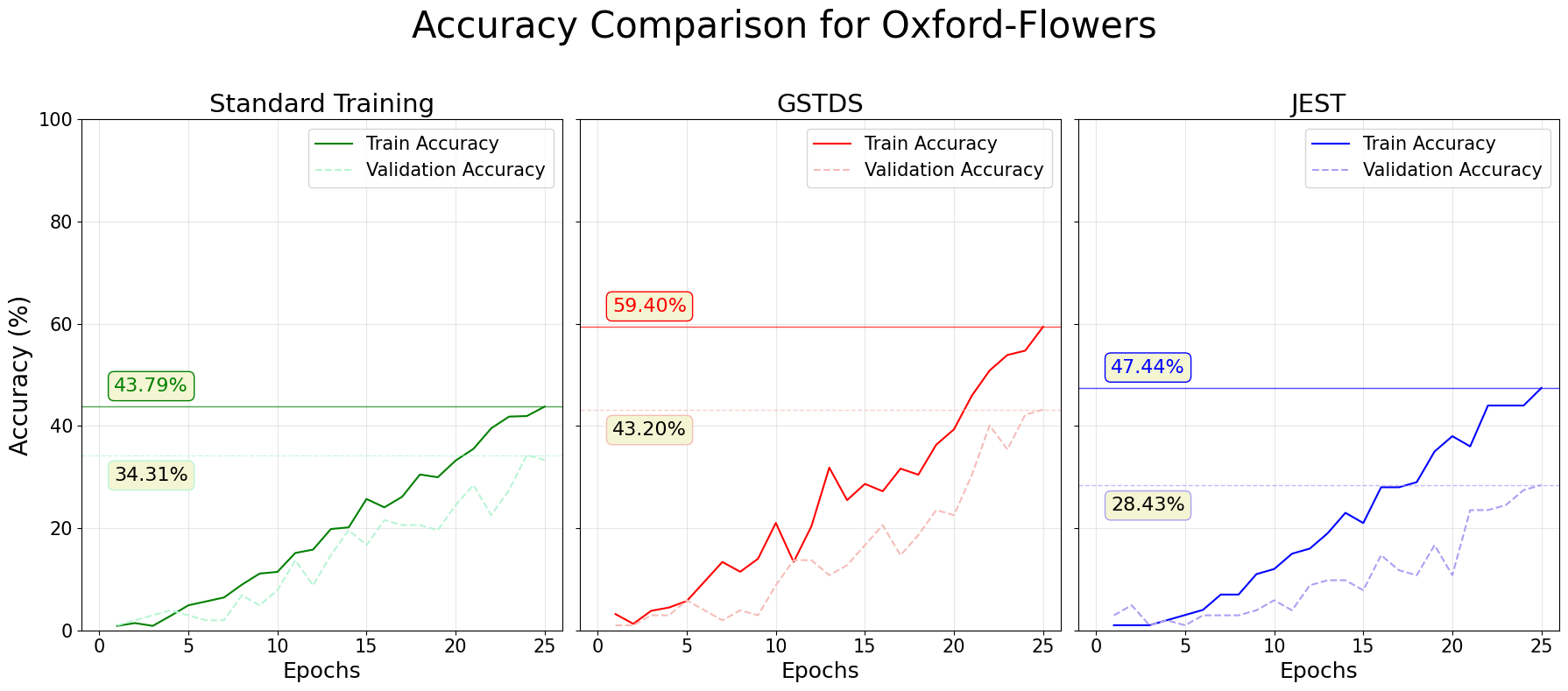}
    \caption{Train and Validation Curves of Oxford-Flowers. From left to right: Standard Training, GSTDS, JEST.}\label{fig:flowers_curves}
\end{figure}

We evaluated the effectiveness of the GSTDS algorithm by comparing it with the JEST algorithm and standard training across multiple datasets. The results are presented in detail for the Oxford-IIIT Pet dataset in Table~\ref{tab:performance_comparison_oxford} and Figure~\ref{fig:Oxford_curves}, for the Oxford-Flowers dataset in Table~\ref{tab:performance_comparison_flowers} and Figure~\ref{fig:flowers_curves}, and for the CIFAR dataset in Table~\ref{tab:performance_comparison_cifar} and Figure~\ref{fig:CIFAR_curves}. These comparisons are based on the adjusted-Sigmoid filter-ratio policy and provide a comprehensive evaluation of the algorithms' performance.

\begin{table}[htbp]
    \caption{Performance Comparison of Different Methods on Oxford-IIIT Pet over 25 epochs}\label{tab:performance_comparison_oxford}
    \begin{tabular*}{\textwidth}{@{\extracolsep\fill}lccc@{}}
    \toprule
    Metric                   & Standard Training & GSTDS          & JEST          \\
    \midrule
    Training Accuracy (\%)   & \textbf{89.82}    & \underline{87.78} & 86.90       \\
    Validation Accuracy (\%) & 84.51             & \textbf{85.87} & \underline{85.33} \\
    Test Accuracy (\%)       & \underline{89.35}             & \textbf{89.69} & 89.12 \\
    Training Loss            & \textbf{0.4490}   & \underline{0.4873} & 0.4922     \\
    Validation Loss          & 0.5617   & \textbf{0.5501}       & \underline{0.5578} \\
    Test Loss                & \underline{0.3501} & \textbf{0.3429} & 0.3585     \\
    Total FLOPs ($10^{13}$)  & 30.1              & \textbf{8.18}  & \underline{15.1} \\
    \botrule
    \end{tabular*}
\end{table}

Table \ref{tab:performance_comparison_oxford} demonstrates that the GSTDS algorithm achieves high accuracy with significantly fewer computational resources, as measured by FLOPs, compared to other methods. This is supported by the convergence behavior illustrated in Figure \ref{fig:Oxford_curves}.




The second dataset is CIFAR-10, which is known for its low-resolution images and significant intra-class variability. The performance of Standard Training, GSTDS, and JEST on this dataset after 25 epochs is quantitatively presented in Table~\ref{tab:performance_comparison_cifar}. Furthermore, the convergence behavior of these methodologies is illustrated in Figure~\ref{fig:CIFAR_curves}.

\begin{table}[htbp]
    \caption{Performance Comparison of Different Methods on CIFAR-10 Pet over 25 epochs}\label{tab:performance_comparison_cifar}
    \begin{tabular*}{\textwidth}{@{\extracolsep\fill}lccc@{}}
    \toprule
    Metric                   & Standard Training & GSTDS          & JEST          \\
    \midrule
    Training Accuracy (\%)   & \underline{79.62} & \textbf{80.16} & 72.81       \\
    Validation Accuracy (\%) & \textbf{81.64}    & \underline{81.34} & 76.86       \\
    Test Accuracy (\%)       & \textbf{82.56}    & \underline{81.89} & 77.72       \\
    Training Loss            & \textbf{0.5055}   & \underline{0.5626} & 0.7832      \\
    Validation Loss          & \textbf{0.4664}   & \underline{0.5073} & 0.6732      \\
    Test Loss                & \textbf{0.5002}   & \underline{0.5203} & 0.6416      \\
    Total FLOPs ($10^{13}$)  & 9.27              & \underline{2.92}  & \textbf{1.45} \\
    \botrule
    \end{tabular*}
\end{table}

As evident from Table~\ref{tab:performance_comparison_cifar}, GSTDS achieves a significant reduction in total computational cost (approximately 70\%) compared to Standard Training, while maintaining a statistically comparable test accuracy. As depicted in Figure \ref{fig:CIFAR_curves}, GSTDS demonstrates a sustained learning trajectory, evidenced by the continued positive slope of its convergence curve at 25 epochs, a behavior not observed in the plateauing curves of JEST and Standard Training. Conversely, JEST, which offers the lowest computational cost (Table~\ref{tab:performance_comparison_cifar}), converges to a lower validation accuracy and falls short of the performance achieved by both Standard Training and GSTDS.

The third dataset, which is Oxford Flowers, presents a significant challenge due to its extensive number of classes and limited instances per class, necessitating effective capture of subtle, fine-grained visual distinctions.

\begin{table}[htbp]
    \caption{Performance Comparison of Different Methods on Oxford-Flowers Pet over 25 epochs}\label{tab:performance_comparison_flowers}
    \begin{tabular*}{\textwidth}{@{\extracolsep\fill}lccc@{}}
    \toprule
    Metric                   & Standard Training & GSTDS          & JEST          \\
    \midrule
    Training Accuracy (\%)   & 43.79             & \textbf{59.40} & \underline{47.44}       \\
    Validation Accuracy (\%) & \underline{34.31} & \textbf{43.20} & 28.43       \\
    Test Accuracy (\%)       & 41.14             & \textbf{58.13} & \underline{45.21}       \\
    Training Loss            & 2.9418            & \textbf{2.3071} & \underline{2.7782}  \\
    Validation Loss          & \underline{3.2850} & \textbf{2.8257} & 3.4279  \\
    Test Loss                & 2.8938            & \textbf{2.1139} & \underline{2.8483}  \\
    Total FLOPs ($10^{13}$)  & 8.34              & \textbf{2.12}   & \underline{3.05}  \\
    \botrule
    \end{tabular*}
\end{table}

As presented in Table~\ref{tab:performance_comparison_flowers}, the GSTDS algorithm significantly outperforms both Standard Training and JEST on this dataset.  GSTDS achieves a superior test accuracy and exhibits the lowest test loss, indicative of enhanced generalization.  Notably, GSTDS attains these performance improvements with reduced computational demands. This efficiency underscores GSTDS's capacity to deliver high accuracy while minimizing computational overhead.  The convergence characteristics, shown in Figure~\ref{fig:flowers_curves}, reveal that JEST displays a wider gap between training and validation losses, which may suggest overfitting tendencies compared to GSTDS.

\subsection{Accuracy Comparison at Fixed FLOPs Number}
\label{subsection:same_flops_accuracy_datasets}

\begin{figure}[htbp]
    \centering
    \includegraphics[width=\textwidth]{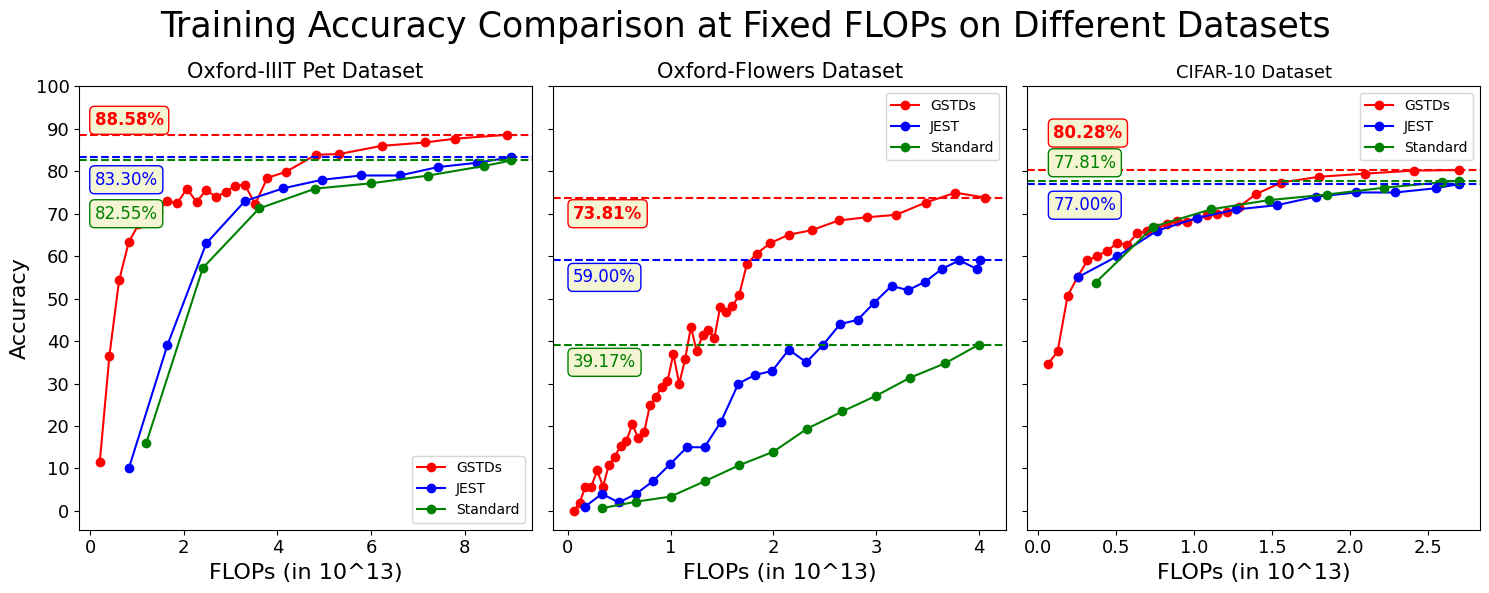} 
    \caption{Accuracy comparison of various training methods across Oxford IIIT Pet, Oxford Flowers, and CFAR-10 datasets, under same FLOPs. Datasets are presented from left to right: Oxford IIIT Pet, Oxford Flowers, and CFAR-10.}
    \label{fig:same_flops_accuracy_datasets}
\end{figure}

Figure~\ref{fig:same_flops_accuracy_datasets} presents a comparative analysis of training method accuracy across Oxford IIIT Pet, Oxford Flowers, and CFAR-10 datasets, evaluated at equivalent FLOPs. 

\begin{table}[htbp]
    \caption{Accuracy Comparison Across Datasets}\label{tab:performance_comparison}
    \centering
    \begin{tabular}{lccccc}
    \toprule
    Dataset & FLOPs & Method & Train Acc. (\%) & Val Acc. (\%) & Test Acc. (\%) \\
    \midrule
    \multirow{3}{*}{Oxford-IIIT-Pet} 
        & \multirow{3}{*}{9$\times 10^{13}$} & Standard & 82.55 & 78.26 & 86.51 \\
        & & GSTDS    & \textbf{88.58} & \textbf{86.24} & \textbf{89.64} \\
        & & JEST     & \underline{83.30} & \underline{79.89} & \underline{87.51} \\
    \midrule
    \multirow{3}{*}{Oxford-Flowers} 
        & \multirow{3}{*}{4$\times 10^{13}$} & Standard & 39.17 & 32.35 & 36.59 \\
        & & GSTDS    & \textbf{73.81} & \textbf{65.36} & \textbf{71.34} \\
        & & JEST     & \underline{59.00} & \underline{42.16} & \underline{56.36} \\
    \midrule
    \multirow{3}{*}{CIFAR-10} 
        & \multirow{3}{*}{2.7$\times 10^{13}$} & Standard & \underline{77.81} & \textbf{80.48} & \textbf{80.98} \\
        & & GSTDS    & \textbf{80.28} & 78.82 & \underline{80.42} \\
        & & JEST     & 77.00 & \underline{80.38} & 79.63 \\
    \bottomrule
    \end{tabular}
    
\end{table}

Figure~\ref{fig:same_flops_accuracy_datasets} illustrates that for both Oxford IIIT Pet and Oxford Flowers datasets, GSTDS achieves notably higher accuracy ($\approx 87.50\%$) compared to other methods under similar computational constraints. While the performance advantage on CFAR-10 is visually less pronounced, GSTDS still gains good accuracy.  This consistent trend across datasets underscores the efficiency and robustness of GSTDS in achieving enhanced performance within resource-limited scenarios.

\subsection{GSTDS Data Selection and Weight Distribution within SVM Regions}
\label{subsection:gstds_data_selection_weight}

\begin{figure}[htbp]
    \centering
    \includegraphics[width=\textwidth]{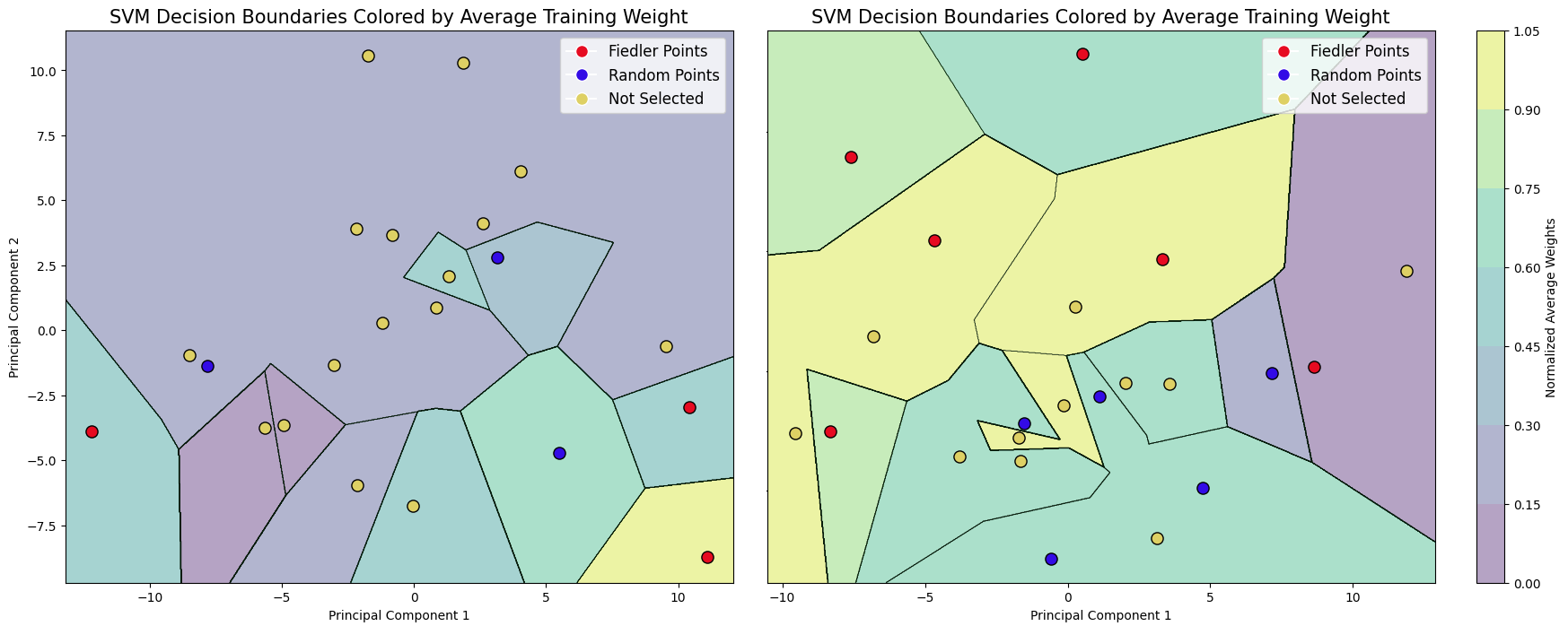}
    \caption{Visualization of GSTDS data selection and weight distribution for Batch 14 of the Oxford Flowers dataset. Left: Epoch~2 with a filter ratio of \(\approx 0.2\). Right: Epoch~31 with a filter ratio of \(\approx 0.8\). }
    \label{fig:gstds_data_selection_weight}
\end{figure}

Figure~\ref{fig:gstds_data_selection_weight} provides a two-dimensional PCA-based visualization of the last batch at an early epoch and a late epoch. In both subplots, data points are colored based on their selection strategy. 
The color of each region represents the average weight gained by the data points until the end of the training process. For each region in the selected batch, we computed the average weights of data points classified under the same SVM decision boundary. By normalizing the corresponding weights, we implicitly defined an informativeness metric. Since Fiedler-based data selection is a highly geometry-sensitive heuristic, we can demonstrate that GSTDS not only selects boundary-critical data points but also those that have accumulated higher average weights by the end of the training process.

Figure~\ref{fig:gstds_data_selection_weight} suggests that, regardless of whether individual data points are specifically selected by GSTDS (via Fiedler vector or random sampling), the overall selection strategy contributes to a beneficial outcome.  This outcome is likely enhanced by the algorithm's tendency to prioritize higher-weight data points while adhering to geometrical criteria defined by the SVM.

\subsection{Analyzing the Contribution of Different Classes in GSTDS Selection Strategy}
\label{subsection:gstds_label_selection_bias}

\begin{figure}[htbp]
    \centering
    \includegraphics[width=0.7\textwidth]{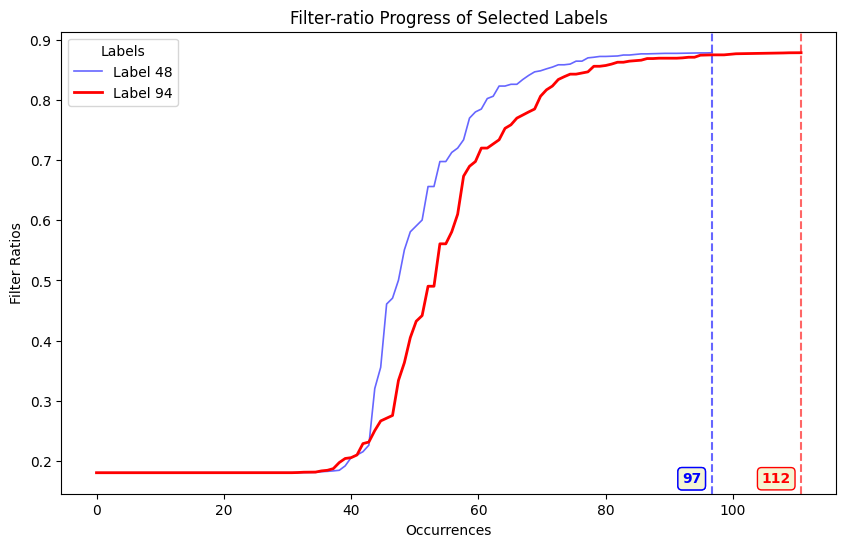}
    \caption{Filter-ratio increase and selection frequency for Labels 94 and 48 during GSTDS training.}
    \label{fig:fr_progress}
\end{figure}

Figure~\ref{fig:fr_progress} illustrates the progression of filter ratios and the corresponding selection frequency for classes with labels 94 and 48 in the Oxford Flowers dataset, demonstrating a wise approach to selecting more challenging data using GSTDS.

\begin{figure}[htbp]
    \centering
    \includegraphics[width=\textwidth]{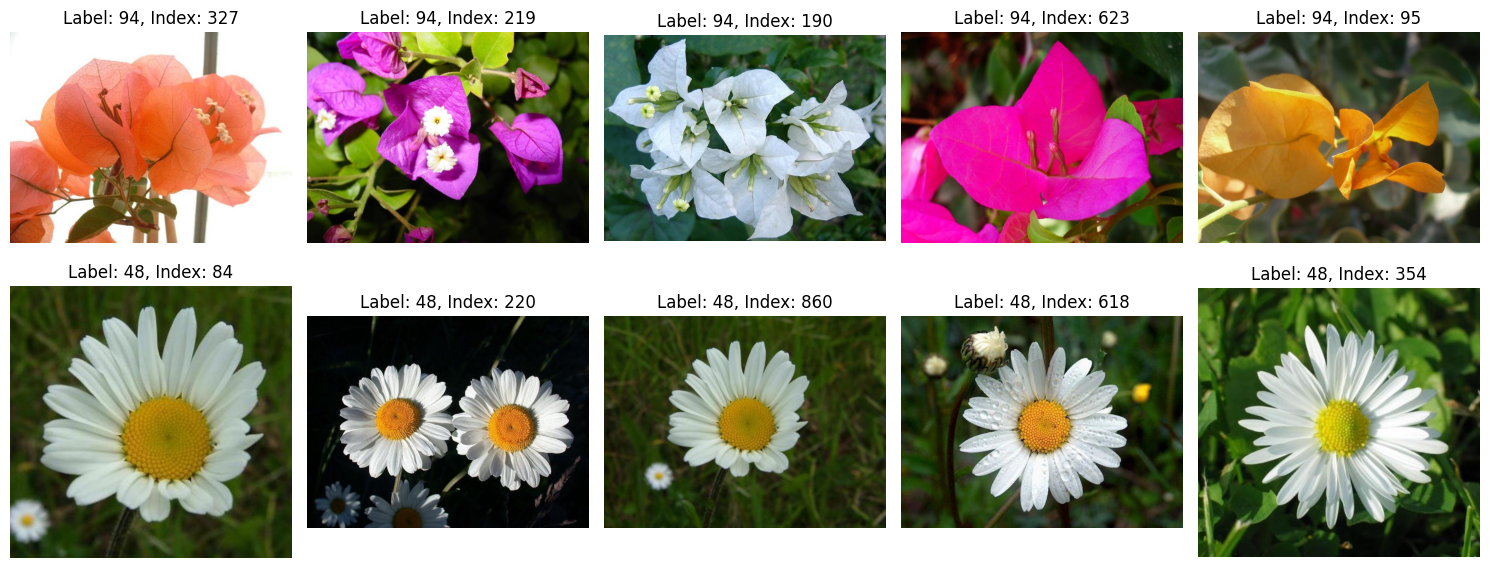}
    \caption{Example images illustrating the intra-class diversity of Label 94 compared to Label 48.}
    \label{fig:label_examples}
\end{figure}

Figure~\ref{fig:label_examples} presents example images of Label 94 compared to Label 48. These observations imply that GSTDS incorporates an adaptive mechanism that prioritizes data points based on their perceived difficulty, thus focusing the learning process on more complex categories.

As shown in Figure~\ref{fig:fr_progress}, Label 94 shows a higher overall selection frequency and filter ratios compared to Label 48 throughout the training process. Specifically, the selection count for Label 94 peaks at 112 occurrences, whereas Label 48 remains around 97. This disparity suggests that GSTDS preferentially samples data from Label 94, potentially because it represents a more challenging or less easily learned category due to its higher intraclass diversity.

\section{Conclusion}\label{conclusion}

We propose \textbf{GSTDS}, a novel data selection algorithm that combines spectral analysis with a curriculum-based filtering schedule to optimize training efficiency and performance in deep learning models. By prioritizing the most informative data points during training, GSTDS enhances generalization and robustness while significantly reducing computational requirements. Experiments across three datasets demonstrate that GSTDS surpasses standard training and the JEST algorithm in both accuracy and efficiency. Its dynamic filtering mechanism and spectral-based data selection facilitate improved convergence and the ability to model complex patterns, establishing GSTDS as an effective approach for training deep learning models across diverse applications.

\section{Future Work}\label{future}
There are too many ideas that need attention in the future work.
 Some of them are listed below:
\begin{itemize}
    \item \textbf{Dynamic Curriculum Learning}: Investigating more advanced curriculum learning strategies to adaptively adjust the filter-ratio sequence based on the model's learning progress and performance.
    \item \textbf{Reinforcement Learning}: Exploring reinforcement learning techniques to dynamically adjust the filter-ratio sequence based on the model's performance on data points and only spend computational resources on the most needed ones (Just learn what you need).

\end{itemize}

\section*{Statements and Declarations}\label{SAD}

\begin{itemize}
    \item \textbf{Competing Interests}:
    The authors declare that they have no known competing financial or non-financial interests that could have appeared to influence the work reported in this paper.
    \item \textbf{Funding}:
    This research received no external funding. The authors have personally financed all aspects of the research.
\end{itemize}
\newpage
 
\bibliography{sn-bibliography}

\end{document}